\documentclass[a4paper,conference]{IEEEtran}
\IEEEoverridecommandlockouts
\usepackage[utf8]{inputenc}
\usepackage[T1]{fontenc}
\usepackage{microtype,newtxtext,newtxmath} 
\usepackage{mathtools,gensymb,eurosym,url,soul} 
\usepackage{algorithm}
\usepackage{algpseudocode}
\usepackage{caption}
\usepackage{subcaption}
\usepackage{longtable,booktabs,array}
\usepackage[backend=biber, sorting=none, style=ieee, maxnames=5]{biblatex}
\usepackage[switch]{lineno}
\usepackage{acronym}
\usepackage{graphicx}
\usepackage{xcolor}
\usepackage{xr} 
\usepackage{comment}
\usepackage{hyperref}
\hypersetup{colorlinks=true, breaklinks=true, 
  urlcolor=blue, linkcolor=blue,anchorcolor=blue,citecolor=blue,
  hypertexnames=true, final=true, 
  pdfpagemode = UseNone, 
  pdfauthor = {},
  pdftitle = {},   
  pdfsubject = {},
  pdfkeywords = {}
}

\graphicspath{{img/} {figures/} {figures/circuits} {figures/intro_diagrams} {figures/results/} {./}}
\DeclareGraphicsExtensions{.pdf,.png,.jpg,.mps}

\IEEEdisplaynontitleabstractindextext

\DeclareSourcemap{
  \maps[datatype=bibtex]{
    \map[overwrite=true]{
      \step[fieldset=url, null]
      \step[fieldset=doi, null]
      \step[fieldset=eprint, null]
    }
  }
}

\acrodef{ADC}[ADC]{Analog to Digital Converter}
\acrodef{ADEXP}[AdExp-I\&F]{Adaptive-Exponential Integrate and Fire}
\acrodef{ADM}[ADM]{Asynchronous Delta Modulator}
\acrodef{AER}[AER]{Address-Event Representation}
\acrodef{AEX}[AEX]{AER EXtension board}
\acrodef{AE}[AE]{Address-Event}
\acrodef{AFM}[AFM]{Atomic Force Microscope}
\acrodef{AGC}[AGC]{Automatic Gain Control}
\acrodef{AI}[AI]{Artificial Intelligence}
\acrodef{AMDA}[AMDA]{AER Motherboard with D/A converters}
\acrodef{ANN}[ANN]{Artificial Neural Network}
\acrodef{API}[API]{Application Programming Interface}
\acrodef{APMOM}[APMOM]{Alternate Polarity Metal On Metal}
\acrodef{ARM}[ARM]{Advanced RISC Machine}
\acrodef{ASIC}[ASIC]{Application Specific Integrated Circuit}
\acrodef{AdExp}[AdExp-IF]{Adaptive Exponential Integrate-and-Fire}
\acrodef{BCM}[BMC]{Bienenstock-Cooper-Munro}
\acrodef{BD}[BD]{Bundled Data}
\acrodef{BEOL}[BEOL]{Back-end of Line}
\acrodef{BG}[BG]{Bias Generator}
\acrodef{BMI}[BMI]{Brain-Machince Interface}
\acrodef{BTB}[BTB]{band-to-band tunnelling}
\acrodef{CAD}[CAD]{Computer Aided Design}
\acrodef{CAM}[CAM]{Content Addressable Memory}
\acrodef{CAVIAR}[CAVIAR]{Convolution AER Vision Architecture for Real-Time}
\acrodef{CA}[CA]{Cortical Automaton}
\acrodef{CCN}[CCN]{Cooperative and Competitive Network}
\acrodef{CDR}[CDR]{Clock-Data Recovery}
\acrodef{CFC}[CFC]{Current to Frequency Converter}
\acrodef{CHP}[CHP]{Communicating Hardware Processes}
\acrodef{CMIM}[CMIM]{Metal-insulator-metal Capacitor}
\acrodef{CML}[CML]{Current Mode Logic}
\acrodef{CMOL}[CMOL]{Hybrid CMOS nanoelectronic circuits}
\acrodef{CMOS}[CMOS]{Complementary Metal-Oxide-Semiconductor}
\acrodef{CNN}[CNN]{Convolutional Neural Network}
\acrodef{COTS}[COTS]{Commercial Off-The-Shelf}
\acrodef{CPG}[CPG]{Central Pattern Generator}
\acrodef{CPLD}[CPLD]{Complex Programmable Logic Device}
\acrodef{CPU}[CPU]{Central Processing Unit}
\acrodef{CSM}[CSM]{Cortical State Machine}
\acrodef{CSP}[CSP]{Constraint Satisfaction Problem}
\acrodef{CTXCTL}[CTXCTL]{CortexControl}
\acrodef{CV}[CV]{Coefficient of Variation}
\acrodef{DAC}[DAC]{Digital to Analog Converter}
\acrodef{DAS}[DAS]{Dynamic Auditory Sensor}
\acrodef{DAVIS}[DAVIS]{Dynamic and Active Pixel Vision Sensor}
\acrodef{DBN}[DBN]{Deep Belief Network}
\acrodef{DFA}[DFA]{Deterministic Finite Automaton}
\acrodef{DIBL}[DIBL]{drain-induced-barrier-lowering}
\acrodef{DI}[DI]{delay insensitive}
\acrodef{DMA}[DMA]{Direct Memory Access}
\acrodef{DNF}[DNF]{Dynamic Neural Field}
\acrodef{DNN}[DNN]{Deep Neural Network}
\acrodef{DOF}[DOF]{Degrees of Freedom}
\acrodef{DPE}[DPE]{Dynamic Parameter Estimation}
\acrodef{DPI}[DPI]{Differential Pair Integrator}
\acrodef{DRAM}[DRAM]{Dynamic Random Access Memory}
\acrodef{DRRZ}[DR-RZ]{Dual-Rail Return-to-Zero}
\acrodef{DR}[DR]{Dual Rail}
\acrodef{DSP}[DSP]{Digital Signal Processor}
\acrodef{DVS}[DVS]{Dynamic Vision Sensor}
\acrodef{DYNAP}[DYNAP]{Dynamic Neuromorphic Asynchronous Processor}
\acrodef{EBL}[EBL]{Electron Beam Lithography}
\acrodef{EDVAC}[EDVAC]{Electronic Discrete Variable Automatic Computer}
\acrodef{EEG}[EEG]{electroencephalography}
\acrodef{EIN}[EIN]{Excitatory-Inhibitory Network}
\acrodef{EM}[EM]{Expectation Maximization}
\acrodef{EPSC}[EPSC]{Excitatory Post-Synaptic Current}
\acrodef{EPSP}[EPSP]{Excitatory Post-Synaptic Potential}
\acrodef{EZ}[EZ]{Epileptogenic Zone}
\acrodef{FDSOI}[FDSOI]{Fully-Depleted Silicon on Insulator}
\acrodef{FET}[FET]{Field-Effect Transistor}
\acrodef{FFT}[FFT]{Fast Fourier Transform}
\acrodef{FI}[F-I]{Frequency-Current}
\acrodef{FNN}[FNN]{Feed-forward Neural Network}
\acrodef{FPGA}[FPGA]{Field Programmable Gate Array}
\acrodef{FR}[FR]{Fast Ripple}
\acrodef{FSA}[FSA]{Finite State Automaton}
\acrodef{FSM}[FSM]{Finite State Machine}
\acrodef{GIDL}[GIDL]{gate-induced-drain-leakage}
\acrodef{GOPS}[GOPS]{Giga-Operations per Second}
\acrodef{GPU}[GPU]{Graphical Processing Unit}
\acrodef{GT}[GT]{Ground Truth}
\acrodef{GUI}[GUI]{Graphical User Interface}
\acrodef{HAL}[HAL]{Hardware Abstraction Layer}
\acrodef{HFO}[HFO]{High Frequency Oscillation}
\acrodef{HH}[H\&H]{Hodgkin \& Huxley}
\acrodef{HMM}[HMM]{Hidden Markov Model}
\acrodef{HRS}[HRS]{High-Resistive State}
\acrodef{HR}[HR]{Human Readable}
\acrodef{HSE}[HSE]{Handshaking Expansion}
\acrodef{HW}[HW]{Hardware}
\acrodef{ICT}[ICT]{Information and Communication Technology}
\acrodef{IC}[IC]{Integrated Circuit}
\acrodef{IEEG}[iEEG]{intracranial electroencephalography}
\acrodef{IF2DWTA}[IF2DWTA]{Integrate \& Fire 2--Dimensional WTA}
\acrodef{IFSLWTA}[IFSLWTA]{Integrate \& Fire Stop Learning WTA}
\acrodef{IF}[I\&F]{Integrate-and-Fire}
\acrodef{IMU}[IMU]{Inertial Measurement Unit}
\acrodef{INCF}[INCF]{International Neuroinformatics Coordinating Facility}
\acrodef{INI}[INI]{Institute of Neuroinformatics}
\acrodef{IO}[I/O]{Input/Output}
\acrodef{IPSC}[IPSC]{Inhibitory Post-Synaptic Current}
\acrodef{IPSP}[IPSP]{Inhibitory Post-Synaptic Potential}
\acrodef{IP}[IP]{Intellectual Property}
\acrodef{ISI}[ISI]{Inter-Spike Interval}
\acrodef{IoT}[IoT]{Internet of Things}
\acrodef{JFLAP}[JFLAP]{Java - Formal Languages and Automata Package}
\acrodef{LEDR}[LEDR]{Level-Encoded Dual-Rail}
\acrodef{LFP}[LFP]{Local Field Potential}
\acrodef{LIF}[LI\&F]{Leak Integrate-and-Fire}
\acrodef{LLC}[LLC]{Low Leakage Cell}
\acrodef{LNA}[LNA]{Low-Noise Amplifier}
\acrodef{LPF}[LPF]{Low Pass Filter}
\acrodef{LRS}[LRS]{Low-Resistive State}
\acrodef{LSM}[LSM]{Liquid State Machine}
\acrodef{LTD}[LTD]{Long Term Depression}
\acrodef{LTI}[LTI]{Linear Time-Invariant}
\acrodef{LTP}[LTP]{Long Term Potentiation}
\acrodef{LTU}[LTU]{Linear Threshold Unit}
\acrodef{LUT}[LUT]{Look-Up Table}
\acrodef{LVDS}[LVDS]{Low Voltage Differential Signaling}
\acrodef{MCMC}[MCMC]{Markov-Chain Monte Carlo}
\acrodef{MEMS}[MEMS]{Micro Electro Mechanical System}
\acrodef{MFR}[MFR]{Mean Firing Rate}
\acrodef{MIM}[MIM]{Metal Insulator Metal}
\acrodef{ML}[ML]{Machine Learning}
\acrodef{MLP}[MLP]{Multilayer Perceptron}
\acrodef{MOSCAP}[MOSCAP]{Metal Oxide Semiconductor Capacitor}
\acrodef{MOSFET}[MOSFET]{Metal Oxide Semiconductor Field-Effect Transistor}
\acrodef{MOS}[MOS]{Metal Oxide Semiconductor}
\acrodef{MRI}[MRI]{Magnetic Resonance Imaging}
\acrodef{NCS}[NCS]{Neuromorphic Cognitive Systems}
\acrodef{NDFSM}[NDFSM]{Non-deterministic Finite State Machine} 
\acrodef{ND}[ND]{Noise-Driven}
\acrodef{NEF}[NEF]{Neural Engineering Framework}
\acrodef{NHML}[NHML]{Neuromorphic Hardware Mark-up Language}
\acrodef{NIL}[NIL]{Nano-Imprint Lithography}
\acrodef{NMDA}[NMDA]{N-Methyl-D-Aspartate}
\acrodef{NME}[NE]{Neuromorphic Engineering}
\acrodef{NN}[NN]{Neural Network}
\acrodef{NOC}[NoC]{Network-on-Chip}
\acrodef{NRZ}[NRZ]{Non-Return-to-Zero}
\acrodef{NSM}[NSM]{Neural State Machine}
\acrodef{OR}[OR]{Operating Room}
\acrodef{OTA}[OTA]{Operational Transconductance Amplifier}
\acrodef{PC}[PC]{Personal Computer}
\acrodef{PCB}[PCB]{Printed Circuit Board}
\acrodef{PCHB}[PCHB]{Pre-Charge Half-Buffer}
\acrodef{PCM}[PCM]{Phase Change Memory}
\acrodef{PE}[PE]{Phase Encoding}
\acrodef{PFA}[PFA]{Probabilistic Finite Automaton}
\acrodef{PFC}[PFC]{prefrontal cortex}
\acrodef{PFM}[PFM]{Pulse Frequency Modulation}
\acrodef{PR}[PR]{Production Rule}
\acrodef{PSC}[PSC]{Post-Synaptic Current}
\acrodef{PSP}[PSP]{Post-Synaptic Potential}
\acrodef{PSTH}[PSTH]{Peri-Stimulus Time Histogram}
\acrodef{QDI}[QDI]{Quasi Delay Insensitive}
\acrodef{RAM}[RAM]{Random Access Memory}
\acrodef{RA}[RA]{Resected Area}
\acrodef{RDF}[RDF]{random dopant fluctuation}
\acrodef{RELU}[ReLu]{Rectified Linear Unit}
\acrodef{RLS}[RLS]{Recursive Least-Squares}
\acrodef{RMSE}[RMSE]{Root Mean Squared-Error}
\acrodef{RMS}[RMS]{Root Mean Squared}
\acrodef{RNN}[RNN]{Recurrent Neural Networks}
\acrodef{ROLLS}[ROLLS]{Reconfigurable On-Line Learning Spiking}
\acrodef{RRAM}[R-RAM]{Resistive Random Access Memory}
\acrodef{R}[R]{Ripples}
\acrodef{SAC}[SAC]{Selective Attention Chip}
\acrodef{SAT}[SAT]{Boolean Satisfiability Problem}
\acrodef{SCX}[SCX]{Silicon CorteX}
\acrodef{SD}[SD]{Signal-Driven}
\acrodef{SEM}[SEM]{Spike-based Expectation Maximization}
\acrodef{SLAM}[SLAM]{Simultaneous Localization and Mapping}
\acrodef{SNN}[SNN]{Spiking Neural Network}
\acrodef{SNR}[SNR]{Signal to Noise Ratio}
\acrodef{SOC}[SOC]{System-On-Chip}
\acrodef{SOI}[SOI]{Silicon on Insulator}
\acrodef{SOZ}[SOZ]{Seizure Onset Zone}
\acrodef{SP}[SP]{Separation Property}
\acrodef{SRAM}[SRAM]{Static Random Access Memory}
\acrodef{STDP}[STDP]{Spike-Timing Dependent Plasticity}
\acrodef{STD}[STD]{Short-Term Depression}
\acrodef{STP}[STP]{Short-Term Plasticity}
\acrodef{STT-MRAM}[STT-MRAM]{Spin-Transfer Torque Magnetic Random Access Memory}
\acrodef{STT}[STT]{Spin-Transfer Torque}
\acrodef{SW}[SW]{Software}
\acrodef{TCAM}[TCAM]{Ternary Content-Addressable Memory}
\acrodef{TFT}[TFT]{Thin Film Transistor}
\acrodef{TLE}[TLE]{Temporal Lobe Epilepsy}
\acrodef{USB}[USB]{Universal Serial Bus}
\acrodef{VHDL}[VHDL]{VHSIC Hardware Description Language}
\acrodef{VLSI}[VLSI]{Very Large Scale Integration}
\acrodef{VOR}[VOR]{Vestibulo-Ocular Reflex}
\acrodef{WCST}[WCST]{Wisconsin Card Sorting Test}
\acrodef{WTA}[WTA]{Winner-Take-All}
\acrodef{XML}[XML]{eXtensible Mark-up Language}
\acrodef{divmod3}[DIVMOD3]{divisibility of a number by three}
\acrodef{hWTA}[hWTA]{hard Winner-Take-All}
\acrodef{sWTA}[sWTA]{soft Winner-Take-All}

\begin{document}
\title{Neuromorphic analog circuits for robust on-chip always-on learning in spiking neural networks}

\author{\IEEEauthorblockN{
Arianna Rubino\IEEEauthorrefmark{1}, 
Matteo Cartiglia\IEEEauthorrefmark{1},
Melika Payvand,
Giacomo Indiveri}
\IEEEauthorblockA{
  Institute of Neuroinformatics,
  University of Zurich and ETH Zurich\\
  Email: [giacomo|rubinoa|camatteo]@ini.uzh.ch }
\thanks{\IEEEauthorrefmark{1} These authors have contributed equally to this work.
This work has received funding from European Union's Horizon 2020 research and innovation program under grant agreement No 871737 (``BeferroSynaptic''), from the European Research Council (ERC) under grant agreement No 724295 (``NeuroAgents''), the  Swiss  National Science Foundation Sinergia project CRSII5-18O316, and the UZH Candoc fellowship FK-22-084.}
}

%

\maketitle

\begin{abstract}
  Mixed-signal neuromorphic systems represent a promising solution for solving extreme-edge computing tasks without relying on external computing resources.
  Their spiking neural network circuits are optimized for processing sensory data on-line in continuous-time.
  However, their low precision and high variability can severely limit their performance.
  To address this issue and improve their robustness to inhomogeneities and noise in both their internal state variables and external input signals, we designed on-chip learning circuits with short-term analog dynamics and long-term tristate discretization mechanisms.
  An additional hysteretic stop-learning mechanism is included to improve stability and automatically disable weight updates when necessary, to enable continuous always-on learning.
  We designed a spiking neural network with these learning circuits in a prototype chip using a 180\,nm CMOS technology.
  Simulation and silicon measurement results from the prototype chip are presented.
  These circuits enable the construction of large-scale spiking neural networks with online learning capabilities for real-world edge computing tasks.
  
\end{abstract}

\begin{IEEEkeywords}
always-on learning, edge-computing, on-chip learning online, SNN, hysteresis, tristability.
\end{IEEEkeywords}

\section{Introduction}
\label{sec:introduction}
The requirements of artificial intelligence (AI) systems operating at the edge are similar to those that living organisms face to function in daily life.
They need to measure sensory signals in real-time, perform closed-loop interactions with their surroundings, be energy-efficient, and continuously adapt to changes in the environment and in their own internal state.
These requisites are well supported by neuromorphic systems and emerging memory technologies that implement brain-inspired mixed-signal spiking neural network (SNN) architectures~\cite{Chicca_etal14,Moradi_etal18,Thakur_etal18,Xia_Yang19,Benjamin_etal21}.
These types of SNNs operate in a data-driven manner, with an event-based representation that is typically sparse in both space and time.
Since they compute only when data is present, they are very power efficient. 
Similar to the biological neural systems they model, these SNNs are particularly well-suited to processing real-world signals.
They can be designed to operate at the same data-rate of the input streams in real-time by matching the time constants of neural computation with those of the incoming signal dynamics.
However, similar to their biological counterparts, these systems are affected by a high degree of variability and sensitivity to noise.
\begin{figure}
  \centering
  \includegraphics[width=0.35\textwidth]{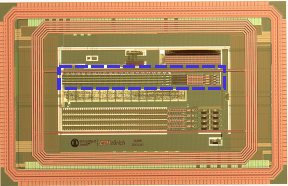}
  \caption{Micrograph of the prototype chip comprising of the learning circuits in a network of 4 neurons with  64 synapses each (see highlighted area). The chip, comprising of additional test structures,  measures $3\times5\,$mm$^2$.}
  \label{fig:chip}
\end{figure}
One of the most effective strategies that biology uses to cope with noise and variability is to utilize adaptation and plasticity.
This strategy has also been adopted by the neuromorphic community: several on-chip implementations of spike-based learning circuits have been proposed in the past~\cite{Hafliger_etal97,Bofill-i-Petit_Murray02,Cartiglia_etal22,Indiveri03b,Chicca_etal03,Mitra_etal09,Mayr_Partzsch10,Bamford_etal12,Qiao_etal15,Lewden_etal20,Uenohara_Aihara22,Payvand_etal22}.
However, few have addressed the problem of being able to operate robustly and autonomously in continuous time, with the ability to switch automatically and reliably between learning and inference modes.
Following the original neuromorphic engineering approach~\cite{Mead20}, we propose a set of analog circuits that faithfully emulate synaptic plasticity mechanisms observed in pyramidal cells of cortical circuits and implement complex spike-based learning and state-dependent mechanisms that support this functionality.
In addition, we extend the concept of long-term bi-stability of synaptic weights, proposed to increase robustness to noise and variability in the input signals~\cite{Senn_Fusi05,Mitra_etal09}, to a tristate stability and weight discretization circuit that increases the resolution of the (stable and crystallized) synaptic weights.
The synaptic plasticity circuits update an internal state variable of the synapse on every pre-synaptic input spike.
The change in this state variable is computed in continuous time by the soma block of the neuron.
In parallel, depending on its value, the internal variable is driven to one of three possible stable states and converted to a discrete three-state synaptic weight current value.
The post-synaptic learning circuits comprise an additional mechanism that gates the weight changes, to stop the learning process when the neuron's mean firing rate is outside a defined learning window.
The circuits were fabricated on a prototype SNN chip designed in a 180\,nm 6M1P CMOS technology and tested within a network of 4 neurons with 64 synapses each (see Fig.~\ref{fig:chip}).
In the following sections we describe the main building blocks used at both the synapse and neuron level, demonstrate their expected behavior with circuit simulations, and provide experimental results measured from the chip. 

\section{Network architecture}
\label{sec:network_architecture}

\begin{figure}
  \centering
  \includegraphics[width=0.35\textwidth]{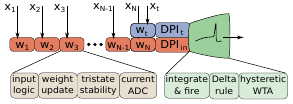}
  \caption{Block diagram of a single neuron row. The plastic synapses (in red) consist of input logic, a weight update, tristate stability, and a current ADC block.
    Input spikes arriving at a synapse update the internal weight variable $\textsf{V}_\textsf{w}$, and change it by an amount that is determined by the post-synaptic learning circuits at the soma (in green).
    The weight voltage is slowly driven to one of three possible stable states, and converted into a synaptic current by thresholding circuits.
    A parallel pathway provides a target current (in blue). Both input and target currents are integrated over time by differential pair integrator (DPI) circuits. The soma (in green), comprises of an Integrate \& Fire (I\&F) block which  integrates the sum of target and input currents, a Delta rule block that calculates the difference of the two DPI currents to determine the amplitude of the weight change, and a hysteretic Winner-Take-All (hWTA) block used to determine if and when to ``stop-learning''.}
   \label{fig:row}
\end{figure}

The block diagram of each neuron in the network is shown in Fig.~\ref{fig:row}.
Input digital events ($\textsf{x}_\textsf{0-N}$) arrive at the individual synapses via asynchronous logic~\cite{Moradi_etal18} and trigger local weight update circuits to induce a change in the voltage stored on a local capacitor by an amount determined by the post-synaptic learning circuits.
In parallel, a tristate stability circuit drives this internal voltage to one of three possible stable states.
This local internal voltage is then discretized and converted to a low, intermediate or high current value.
All currents produced by all synapses are summed spatially and conveyed to a differential pair integrator (DPI), which integrates the weighted sum over time~\cite{Bartolozzi_Indiveri07a}.
A parallel and analogous pathway receives input events representing a desired target signal ($\textsf{x}_\textsf{t}$), and produces a corresponding current from its dedicated DPI circuit. The target and input currents are both summed to drive the neuron's post-synaptic Integrate \& Fire (I\&F) circuit~\cite{Rubino_etal20}, and subtracted to drive the soma's Delta rule circuit~\cite{Payvand_Indiveri19}.
The Delta rule circuit produces either positive or negative weight update signals proportional to the difference between the target input and the weighted synaptic input.
These signals are broadcast to all the neuron's input synapses in continuous time if learning is enabled.
Learning is enabled (or disabled) by means of two hysteretic Winner-Take-All (hWTA) circuits~\cite{Indiveri97a} that compare the neuron's mean output firing rate to a low and a high threshold (see Section~\ref{sec:stop_learning_description} for details).

\section{Circuits}
\label{sec:circuits}

As the details of the Delta rule and I\&F circuits have already been presented~\cite{Payvand_Indiveri19,Rubino_etal20,Cartiglia_etal22}, we describe the synapse learning circuits and the soma hWTA circuit. 

\begin{figure}
  \centering
  \begin{subfigure}{0.12\textwidth}
    \includegraphics[width=\textwidth]{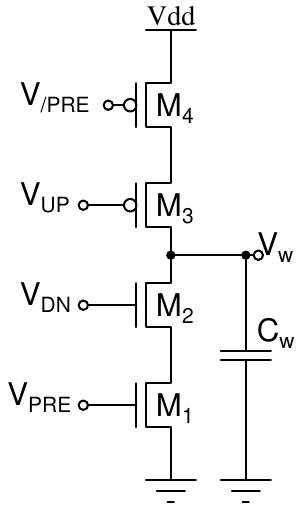}
    \caption{}
    \label{fig:wupdn}
  \end{subfigure}
  \begin{subfigure}{0.18\textwidth}
    \includegraphics[width=\textwidth]{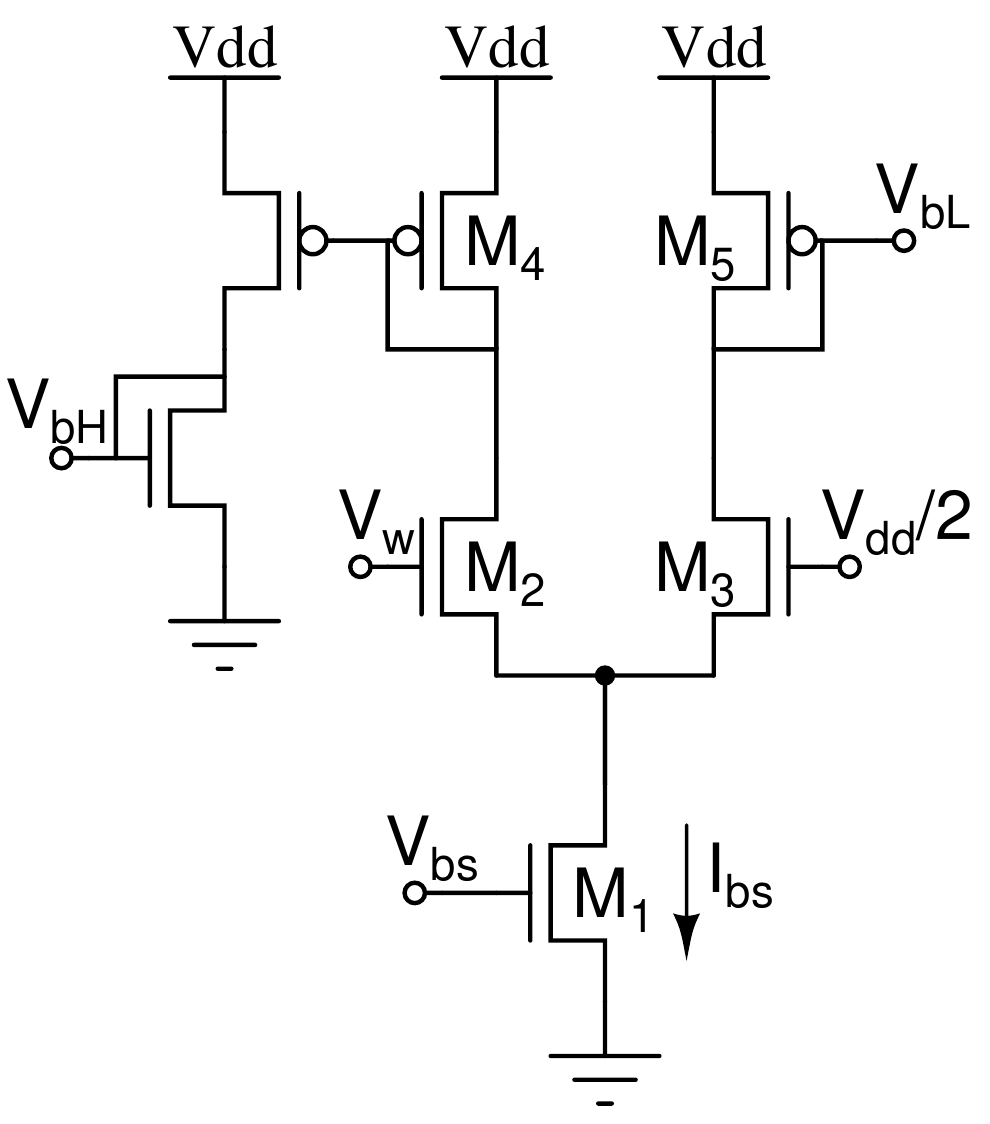}
    \caption{}
    \label{fig:tribias}
  \end{subfigure}
  \begin{subfigure}{0.16\textwidth}
    \includegraphics[width=\textwidth]{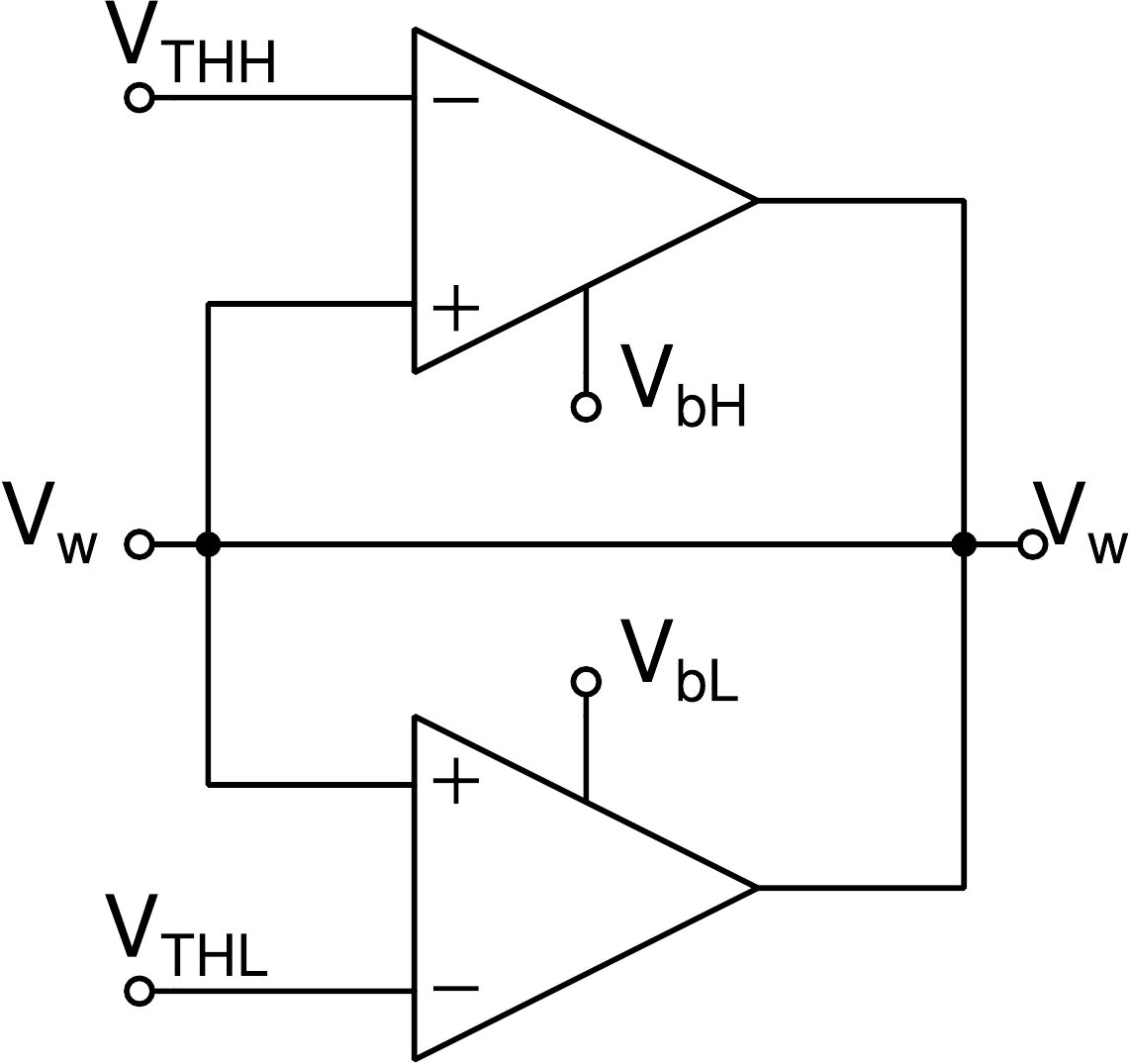}
    \caption{}
    \label{fig:tristability}
  \end{subfigure}\\
  \begin{subfigure}{0.3\textwidth}
    \includegraphics[width=\textwidth]{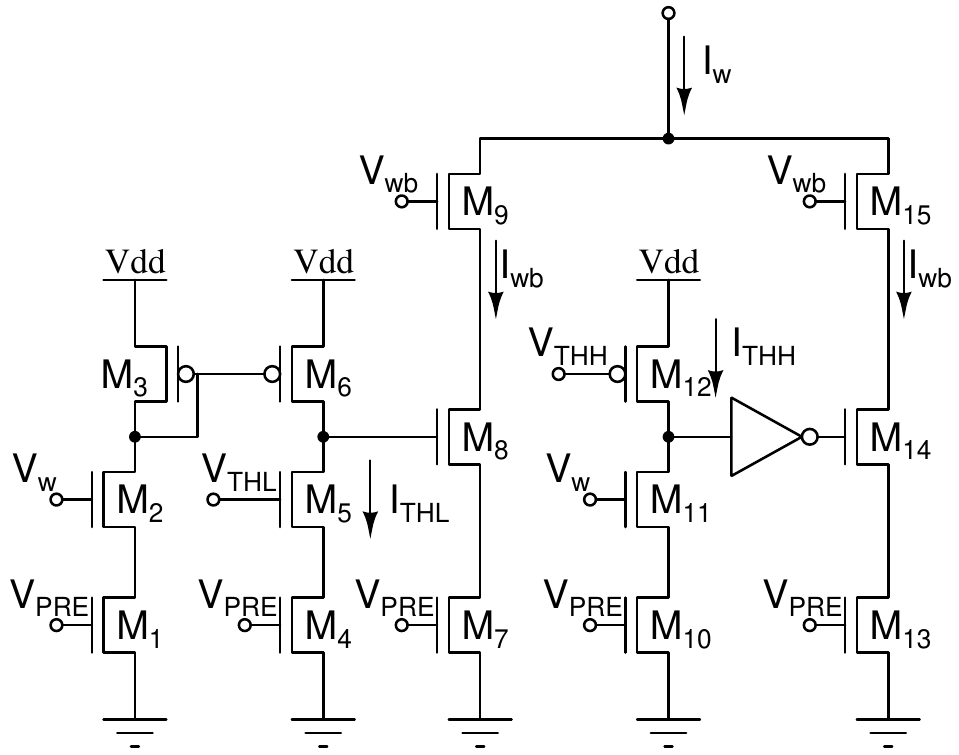}
    \caption{}
    \label{fig:wdiscretization}
  \end{subfigure}  
  \caption{Synapse learning circuits. (\subref{fig:wupdn})~The weight update circuit increases or decreases the internal analog weight variable $\textsf{V}_\textsf{w}$ with every input spike $\textsf{V}_\textsf{PRE}$;  (\subref{fig:tribias})~The tri-stability supply voltage circuit determines which of the two amplifiers in (\subref{fig:tristability}) to power depending on the value of $\textsf{V}_\textsf{w}$ with respect to $\textsf{V}_\textrm{dd}/2$ by activating either  $\textsf{V}_\textsf{bH}$ or $\textsf{V}_\textsf{bL}$.
  (\subref{fig:tristability})~The tristability circuit drives the $\textsf{V}_\textsf{w}$ voltage towards ground, $\textsf{V}_\textrm{dd}/2$ or $\textsf{V}_\textrm{dd}$ depending on its value relative to $\textsf{V}_\textsf{THH}$ and $\textsf{V}_\textsf{THL}$. (\subref{fig:wdiscretization})~The current discretization circuit converts $\textsf{V}_\textsf{w}$ into a low (the leakage current $\textsf{I}_\textsf{0}$), intermediate, or high current.}
  \label{fig:synapse-blocks}
\end{figure}

\subsection{Plastic synapse circuit}
\label{sec:synapse_description}

\begin{figure}
\centering
\includegraphics[width=0.35\textwidth]{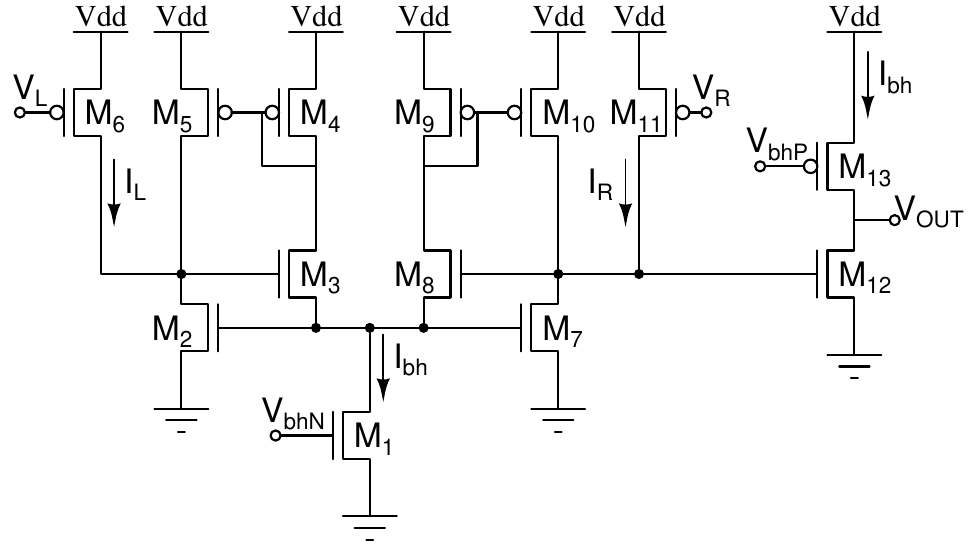}
\caption{hysteresis WTA circuit used to determine the ``stop-learning'' signals. The digital output voltage $\textsf{V}_\textsf{OUT}$ switches from low to high only if the left input current $\textsf{I}_\textsf{L}$ increases above $\textsf{I}_\textsf{R}$ by an amount at least equal to $\textsf{I}_\textsf{bh}$.}
\label{fig:hwta}
\end{figure}

Figure~\ref{fig:synapse-blocks} presents all of the learning circuits used at the synaptic level.
With every pre-synaptic spike, the weight update circuit (Fig.~\ref{fig:wupdn}) increases or decreases the internal analog weight variable $\textsf{V}_\textsf{w}$ by an amount determined by the voltages $\textsf{V}_\textsf{UP}$ and $\textsf{V}_\textsf{DN}$, produced by the post-synaptic Delta rule circuits.
The tri-stability supply voltage circuit (Fig.~\ref{fig:tribias}), produces the biases that power either of the positive feedback amplifiers of Fig.~\ref{fig:tristability}, depending on the state of $\textsf{V}_\textsf{w}$ with respect to $\textsf{V}_\textrm{dd}/2$.
The tri-stability circuit (Fig.~\ref{fig:tristability}) consists of two slew-rate limited positive feedback amplifiers which slowly drive  $\textsf{V}_\textsf{w}$ towards ground, $\textsf{V}_\textrm{dd}/2$, or $\textsf{V}_\textrm{dd}$ depending on the value of $\textsf{V}_\textsf{w}$ relative to $\textsf{V}_\textsf{THH}$ and $\textsf{V}_\textsf{THL}$. 
The weight discretization circuit (Fig.~\ref{fig:wdiscretization}) sets the value of the effective synaptic current $\textsf{I}_\textsf{w}$ to
$\textsf{I}_\textsf{0}$, $\textsf{I}_\textsf{wb}$, or $2\textsf{I}_\textsf{wb}$ depending on the state of $\textsf{V}_\textsf{w}$ with respect to $\textsf{V}_\textsf{THL}$ and $\textsf{V}_\textsf{THH}$.


\subsection{Hysteretic WTA for ``stop-learning''}
\label{sec:stop_learning_description}

Figure~\ref{fig:hwta} shows an instance of a hWTA circuit: it consists of two identical cells, ($M_{2}$--$M_{6}$) and ($M_{7}$--$M_{11}$) that compete with each other.
As soon as one cell wins (e.g., the left one), the bias current $\textsf{I}_\textsf{bh}$ is copied and added to the input current of the winning branch (e.g., $\textsf{I}_\textsf{L}$).
This creates a hysteresis window, such that for the winning (left) cell to lose the competition, its input current has to decrease below the input current of the opposite branch by an additional factor equal to the bias current ($\textsf{I}_\textsf{L}<\textsf{I}_\textsf{R} - \textsf{I}_\textsf{bh}$).
The output voltage of this circuit $\textsf{V}_\textsf{OUT}$ switches to ``high'' when the left cell wins, and to ``low'' when the right cell becomes the winner.

To implement the ``stop-learning'' mechanism~\cite{Mitra_etal09}, we produce a current $\textsf{I}_\textsf{Ca}$ (a surrogate of the neuron's calcium concentration) by integrating the post-synaptic neuron spikes with a DPI circuit~\cite{Bartolozzi_Indiveri07a}.
We then compare this current to two thresholds with the two hWTA circuits.
The digital output nodes of the two hWTA circuits were connected to logic gates to produce an active high $\textsf{Learn}$ signal when the $\textsf{I}_\textsf{Ca}$ current is within the set bounds (i.e., within the learning region) and a low when it is outside this region.
This $\textsf{Learn}$ signal is then used as a ``third factor'' to enable or disable the Delta rule weight circuit, and switch on or off the weight updates.
The hysteresis windows of the hWTA circuits are used to distinguish between cases in which the target input is present (to enable learning) or absent (to disable learning and automatically switch to an ``inference'' mode). The effect of this window is described in Section~\ref{sec:simulation-results}.


\section{Results}
\label{sec:results}

We validate the learning circuits with both circuit simulations
and with experimental results measured from  the fabricated chip.

\begin{figure}
  \centering
  \begin{subfigure}[t]{0.49\textwidth}
    \centering
    \includegraphics[width=1\textwidth]{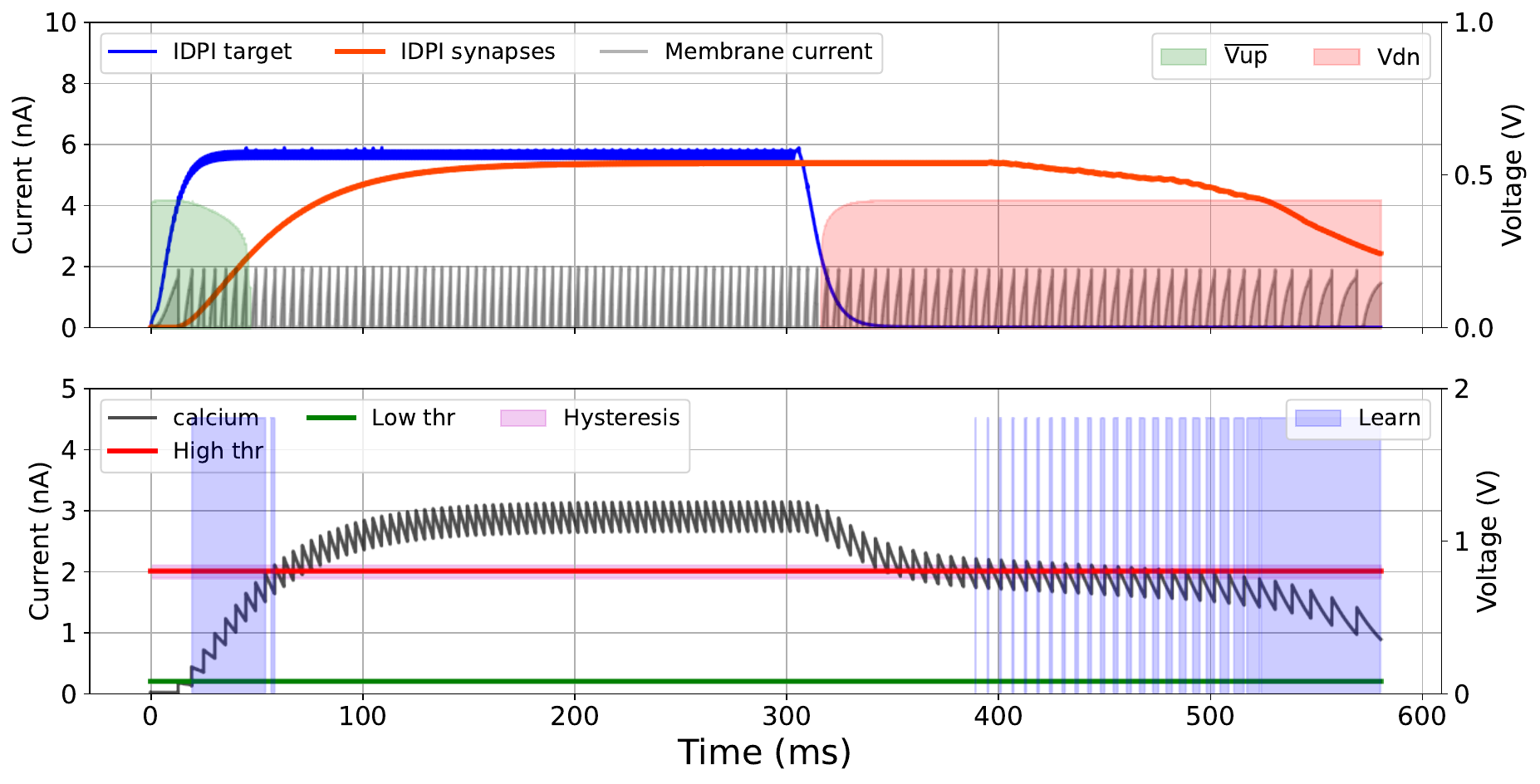}
    \caption{}
    \label{fig:non_hyst}
  \end{subfigure}\\
  \begin{subfigure}[t]{0.49\textwidth}
    \centering
    \includegraphics[width=1\textwidth]{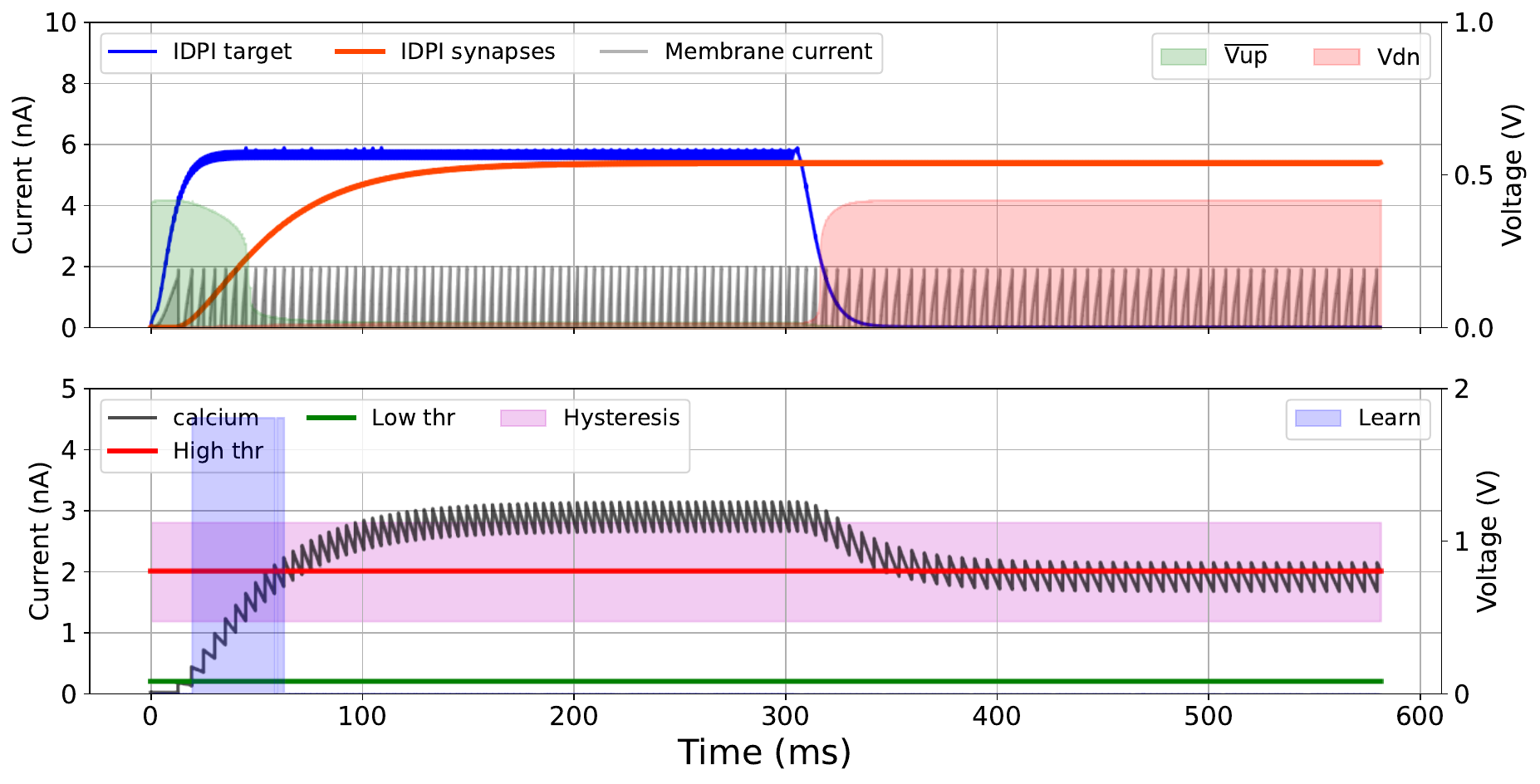}
    \caption{}
    \label{fig:hyst}
  \end{subfigure}
  \caption{Row simulation results. In the top plots of both sub-figures,
    the DPI synapse current (in red) follows the  DPI target current (in blue) when the target input is high. The neuron's membrane activity (in grey) is scaled for visibility. In the lower plots of both sub-figures the calcium current (in black) enables the $\textsf{Learn}$ signal when it lies between the learning region's low (green line) and high (red line) thresholds.
    (\subref{fig:non_hyst}) Small hysteresis bias ($\textsf{I}_\textsf{bh}$ = 100\,pA): after the target current is removed the calcium current drops back into the learning region, the weights are decreased, and the neuron forgets its tuning. (\subref{fig:hyst}) Large hysteresis bias ($\textsf{I}_\textsf{bh}$ = 800\,pA): the large hysteresis region around the highest threshold (shaded pink) keeps the learning disabled, despite the fact that the calcium current fell below the high threshold. The weights of the plastic synapses do not change and the neuron maintains its proper tuning to the trained pattern.}
  \label{fig:hysteresis_simulation}
\end{figure}


\subsection{Circuit simulation results}
\label{sec:simulation-results}
Here, we show simulations of a single neuron and 40 plastic synapses during a learning task and show how the hWTA enables automatic switching from learning to inference.

After initializing all synaptic weights to zero, we started a training phase by stimulating each plastic synapse with a 25\,Hz input spike train, and by sending a spike train with a 1\,kHz frequency to the target synapse.
As expected, during this training phase, the weights of the synapses potentiated and the total weighted synaptic input current increased (see the red trace of Fig.~\ref{fig:hysteresis_simulation}).
During the inference phase, we removed the target input spike train while keeping on stimulating the input synapses.
As expected, without this extra input, the average mean firing rate of the neuron decreased, and the calcium concentration current fell below the upper bound of the learning region.
Figure~\ref{fig:hysteresis_simulation} shows this task performed with two values for $\textsf{I}_\textsf{bh}$, which governs the width of the hysteresis window.
Without a proper hysteresis window (Fig.~\ref{fig:non_hyst}), when the neuron falls back into a learning region it ``forgets'' its training (i.e., the learning circuits decrease the weights).
On the other hand, by properly tuning the hysteresis window (Fig.~\ref{fig:hyst}), the network remains in a ``stop-learning'' mode, and the neuron retains a high output firing rate response to the trained pattern, even in absence of a target signal. In the larger hysteresis window case, the total estimated power consumption is 1.07\,uW, and a maximum (mean) energy of 740\,pJ (680\,pJ) is required to update the weights.

\subsection{Chip measurement results} 
\label{sec:chip measurements}

\begin{figure}
  \centering
    \begin{subfigure}{0.49\textwidth}
        \centering
        \includegraphics[width=1\textwidth]{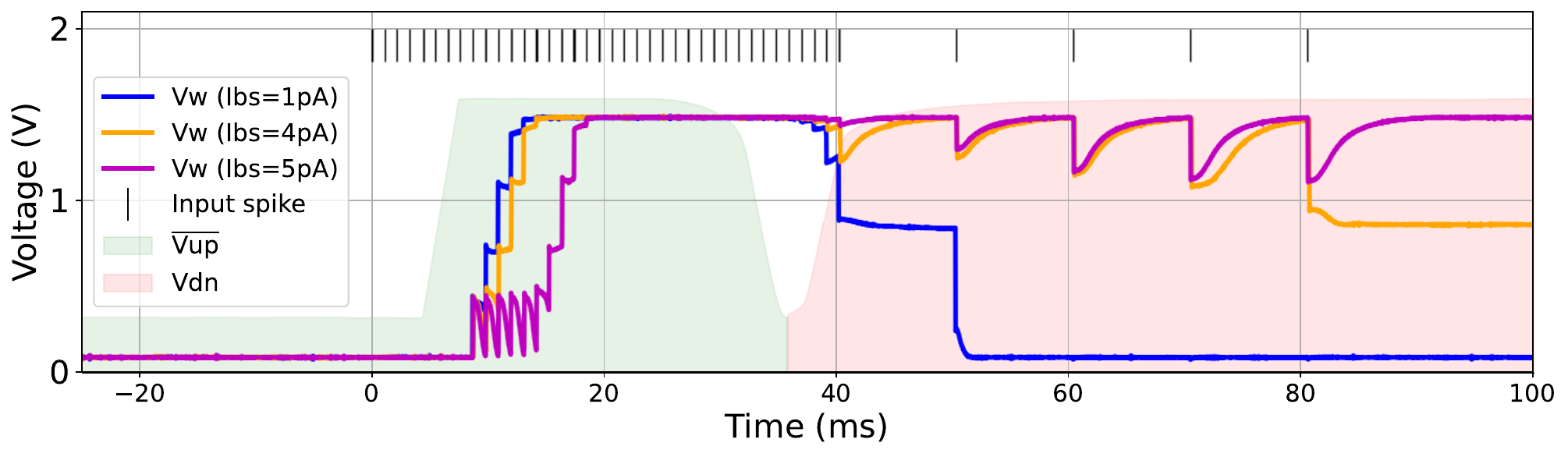}
    \caption{}
    \label{fig:stable_vdd_gnd}
  \end{subfigure}
    \begin{subfigure}{0.49\textwidth}
        \centering
        \includegraphics[width=1\textwidth]{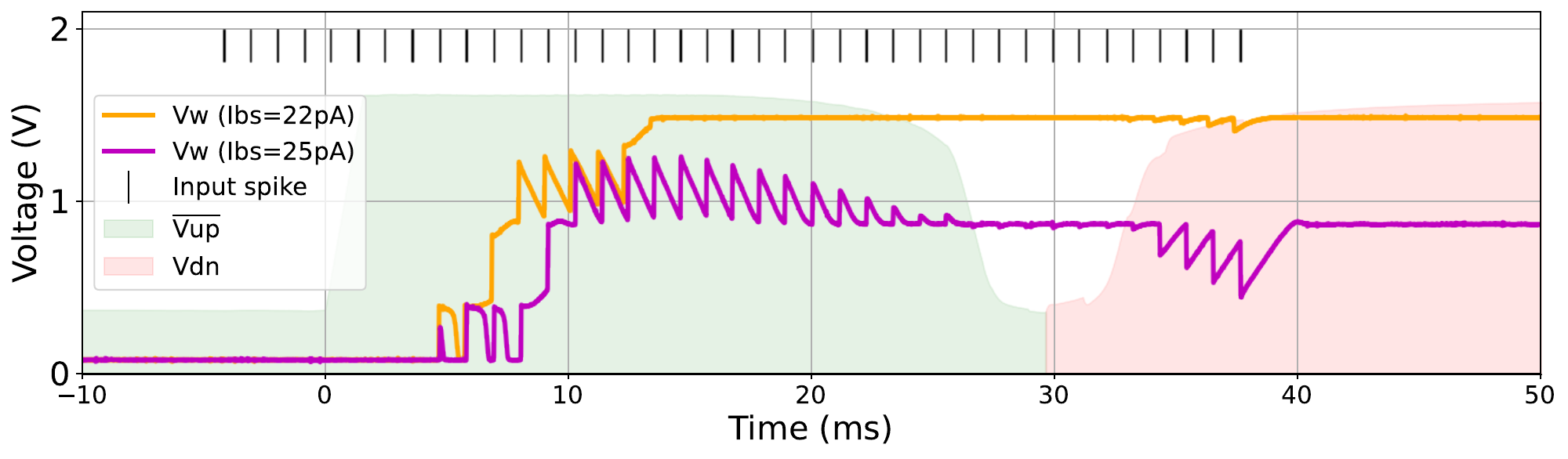}
    \caption{}
    \label{fig:stable_vdd_2}
  \end{subfigure}
    \caption{Tristability chip measurements. (\subref{fig:stable_vdd_gnd}) Stability at $\textsf{V}_\textrm{dd}$ and ground: When the target is presented, the plastic synapse weight is increased by the post-synaptic learning circuits ($\overline{\textsf{V}_\textrm{up}}$ and $\textsf{V}_\textrm{dn}$ scaled for visibility). As the tristability bias increases, the circuit opposes the weight update more strongly. When the target is removed, the tristability maintains the weight value around $\textsf{V}_\textrm{dd}$ for larger values of $\textsf{I}_\textsf{bs}$ (in orange and purple). 
    (\subref{fig:hyst}) Stability at $\textsf{V}_\textrm{dd}/2$: as $\textsf{I}_\textsf{bs}$ increases the tristability opposes the learning with more strength.  
    }
    \label{fig:tristability_results}
  \end{figure}

\subsubsection{Tristability}
\label{sec:tristability_results}

The results from the measurements of the plastic synapse circuits are shown in Fig.~\ref{fig:tristability_results}.
Initially, the neuron is presented with a high target activity, triggering large positive weight updates and causing a rapid increase in the synapse weight internal variable. Upon removal of the target, the weight is decreased.
By increasing the power to the tristate stability amplifiers (Fig.~\ref{fig:tristability}), i.e., by increasing
$\textsf{I}_\textsf{bs}$ of Fig.~\ref{fig:tribias}, the circuit opposes the weight changes more strongly and drives $\textsf{V}_\textsf{w}$ to one of the three stable states more quickly.
Stability at $\textsf{V}_\textrm{dd}$ and ground is shown in Fig.~\ref{fig:stable_vdd_gnd} and at $\textsf{V}_\textrm{dd}/2$ in Fig.~\ref{fig:stable_vdd_2}.
Once the stimulation ends, the tristability circuit crystalizes the weight to one of the three stable states depending on the value of $\textsf{I}_\textsf{bs}$.

\subsubsection{Hysteresis for ``stop-learning''}
\label{sec:stop_learning_results}

\begin{figure}
\centering
\includegraphics[width=0.485\textwidth]{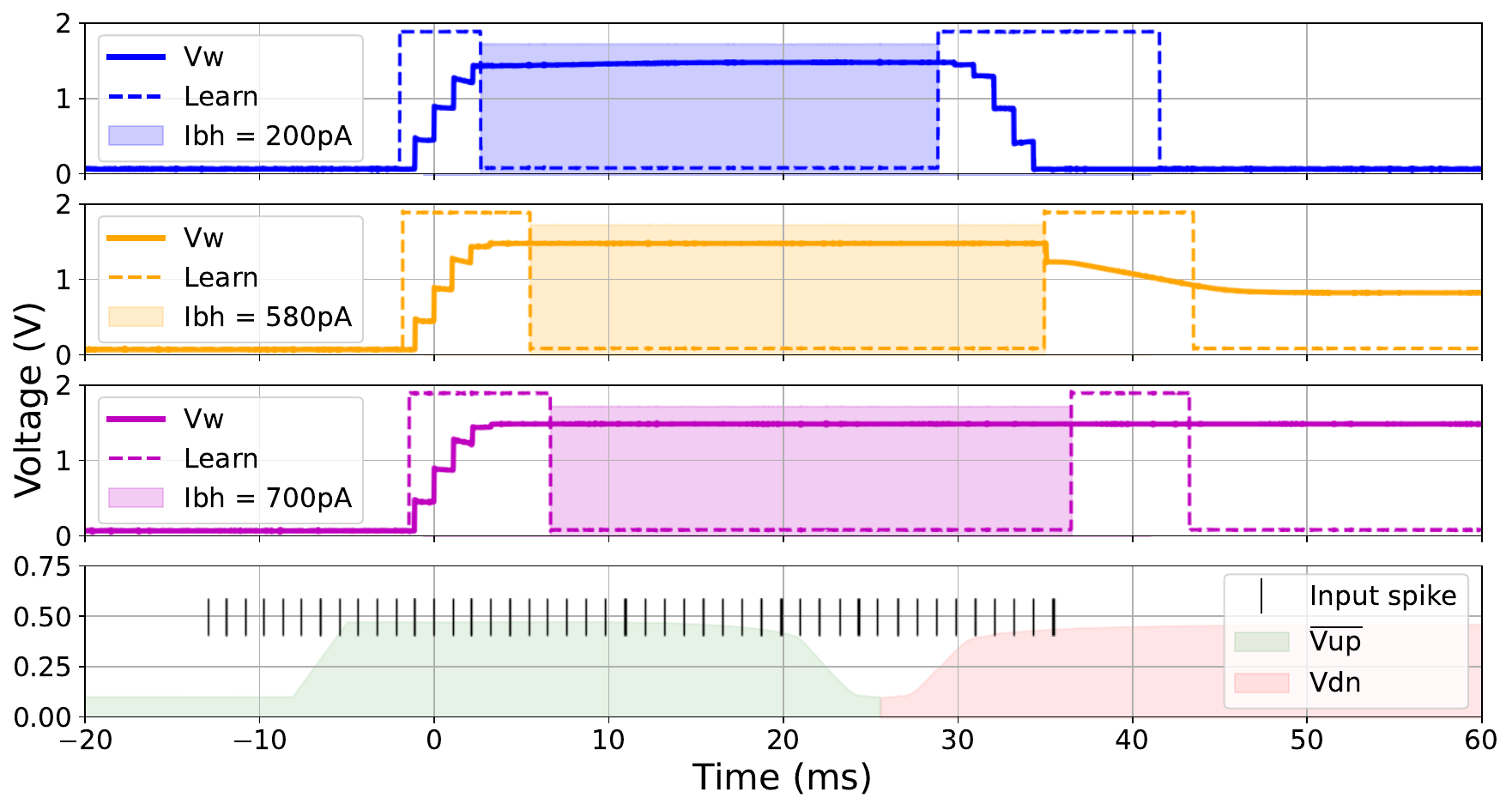}
\caption{Hysteresis chip measurements. The hysteresis window is shown for three different values of $\textsf{I}_\textsf{bh}$. Increases in $\textsf{I}_\textsf{bh}$ produce larger hysteresis windows which are useful for tuning the ``stop-learning'' properties of the network.}
\label{fig:results_hysteresis}
\end{figure}

Figure~\ref{fig:results_hysteresis} shows the results of the characterization of the hysteretic calcium-based stop-learning mechanism. 
Similarly to the previous experiment, the neuron is initially stimulated with a high target activity, bringing it to the learning region. The plastic synapse weight rapidly increases, pushing the neuron into the ``stop-learning'' region. Once the target activity is removed, the neuron returns to the learning region, and, for small hysteresis window settings (top blue plot in Fig.~\ref{fig:results_hysteresis}), the plastic synapse decreases its weight as it is stimulated.
For higher values of $\textsf{I}_\textsf{bh}$ the hysteresis window increases (orange plot in Fig.~\ref{fig:results_hysteresis}) and when the target is removed, the neuron's return to the learning mode is delayed. As this delay increases, even though the plastic synapse keeps on being stimulated, the neuron remains in the ``stop-learning'' region and the weight remains unchanged (purple plot in Fig.~\ref{fig:results_hysteresis}).

\begin{table}
  \caption{Comparison to the state-of-the-art}
  \label{tab:comp-soa} 
\centering
\resizebox{\linewidth}{!}{%
\begin{tabular}{@{}m{0.065\textwidth} m{0.075\textwidth} m{0.075\textwidth} m{0.075\textwidth} m{0.075\textwidth} m{0.08\textwidth}@{}}
  \toprule
  & \textbf{\cite{Frenkel_etal19}}&\textbf{\cite{Davies_etal18}}&\textbf{\cite{Qiao_etal15}}&\textbf{\cite{Cartiglia_etal22}}&\textbf{[This work]}\\
  \midrule
  \textbf{Technology} & 28\,nm & 14\,nm & 180\,nm & 180\,nm & 180\,nm \\
  \textbf{Design} & digital & digital & mixed-signal & mixed-signal & mixed-signal\\
  \textbf{Learning} & semi supervised & program\-mable & semi supervised & error based & semi supervised\\
  \textbf{Stop learning} & yes & no & yes & no & yes \\
  \textbf{Weight resolution} & 3bit & 9bit & bi-stable & 4bit & tri-stable\\
  \textbf{Energy/SOP} & 12.7\,pJ & 120\,pJ & 360\,pJ & 720\,pJ & 680\,pJ\\
  \textbf{Power supply} & 0.55\,V & 0.75\,V & 1.8\,V & 1.8\,V & 1.8\,V\\
  \bottomrule
\end{tabular} }
\end{table}

\section{Conclusions}
\label{sec:conclusion}
We presented a set of analog circuits that enable learning in mixed-signal neuromorphic SNNs, with tristate stability and weight discretization circuits.
By comparing the neuron's calcium concentration to lower and upper bounds, and by using hysteresis, we demonstrate effective always-on learning features, that automatically switch from learning mode to inference mode, without having to manually disable or enable learning. 
Comparisons to previous efforts are provided in Table~\ref{tab:comp-soa}.

\section*{Acknowledgment}
The authors thank Shyam Narayanan, Charlotte Frenkel, and Junren Chen for fruitful discussions and contributions.

\printbibliography

\newpage

\end{document}